\documentclass{article}

     \PassOptionsToPackage{numbers,sort&compress}{natbib}

\usepackage[preprint,nonatbib]{neurips_2019}
\usepackage[square,sort,comma,numbers]{natbib}

\usepackage[utf8]{inputenc} 
\usepackage[T1]{fontenc}    
\usepackage{hyperref}       
\usepackage{url}            
\usepackage{booktabs}       
\usepackage{amsfonts}       
\usepackage{nicefrac}       
\usepackage{microtype}      

\usepackage{graphicx}

\title{Neural Painters: A learned differentiable constraint for generating brushstroke paintings}

\author{
  Reiichiro Nakano\thanks{\url{https://reiinakano.github.io}} \\
  Tokyo, Japan\\
  Independent Researcher\\
  \texttt{reiichiro.s.nakano@gmail.com}
}

\begin{document}

\maketitle

\begin{abstract}
We explore neural painters, a generative model for brushstrokes learned from a real non-differentiable and non-deterministic painting program. We show that when training an agent to “paint” images using brushstrokes, using a differentiable neural painter leads to much faster convergence. We propose a method for encouraging this agent to follow human-like strokes when reconstructing digits. We also explore the use of a neural painter as a differentiable image parameterization. By directly optimizing brushstrokes to activate neurons in a pre-trained convolutional network, we can directly visualize ImageNet categories and generate “ideal” paintings of each class. Finally, we present a new concept called intrinsic style transfer. By minimizing only the content loss from neural style transfer, we allow the artistic medium, in this case, brushstrokes, to naturally dictate the resulting style.\footnote{We have provided Google Colaboratory notebooks to fully reproduce our experiments from scratch at \url{https://github.com/reiinakano/neural-painters/tree/master/notebooks}}
\end{abstract}

\section{Introduction}

\begin{figure}[!htb]
  \centering
  \includegraphics[width=\textwidth, keepaspectratio]{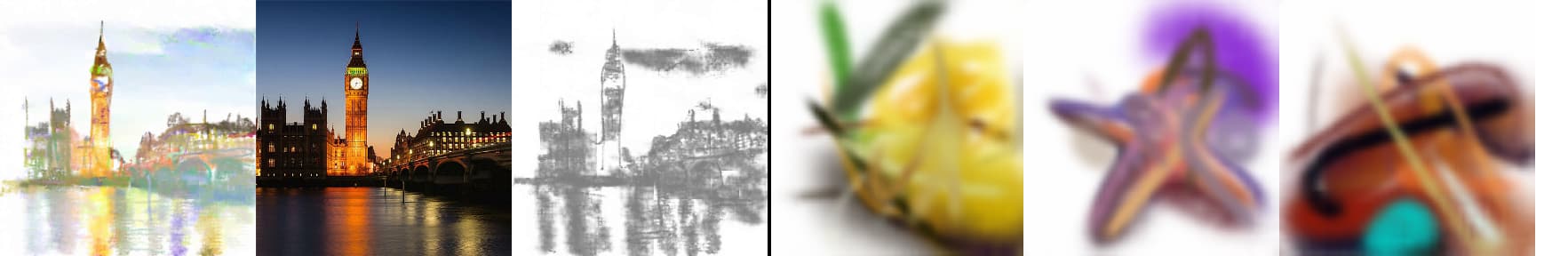}
  \caption{Using a neural painter as a differentiable image parameterization \cite{mordvintsev2018differentiable}, we are able to directly optimize brushstrokes to minimize style transfer's content loss (left), or visualize ImageNet classes (right; pineapple, starfish, and violin categories).}
  \label{fig:topfig}
\end{figure}

There has been much work on using neural networks to generate painting-like images, some of the most notable being style transfer
\cite{gatys2015} and GANs \cite{goodfellow2014:adversarial}, and all their many variations\cite{elgammal2017can, jing2017neural, Gatys_2016_CVPR}. Most of these techniques generate images by having a network directly calculate the RGB value of each pixel.

However, artists don't paint by generating each individual pixel, they paint by generating \textit{brushstrokes}.

As early as 1990~\cite{haeberli1990paint}, long before the sudden popularity of deep learning~\cite{lecun2015deep}, there has been extensive work on automatically finding a set of brushstrokes to “paint” images~\cite{haeberli1990paint, shiraishi2000algorithm, hertzmann1998painterly, winkenbach1994computer}. \textit{Stroke-based rendering} (SBR) is the field of digitally generating paintings by arranging brushstrokes on a canvas according to some optimization goal~\cite{hertzmann2003survey}. The simple idea of using actual brushstrokes gives the outputs of SBR algorithms a very real painting-like texture.

Recently, there has been significant progress on getting neural networks to produce paintings by generating brushstrokes \cite{xie2013artist, ganin2018synthesizing, frans2018unsupervised} instead of pixels. An example is the work done by \citet{ganin2018synthesizing} on SPIRAL, a reinforcement learning agent trained adversarially to learn how to use a real painting program to generate images. A neural agent called SPIRAL learns to reconstruct images from MNIST, Omniglot, and CelebA by defining brushstrokes on a canvas.

Though impressive, the reinforcement adversarial learning framework used by SPIRAL is complex and requires significant computational resources~\cite{ganin2018synthesizing}. Recent work by \citet{ha2018recurrent} and \citet{hafner2018learning} have shown that building a differentiable world model of an environment can drastically reduce computational requirements for reinforcement learning algorithms.

In this paper, we perform various experiments with neural painters, which are differentiable simulations of a non-differentiable painting program. Our contributions are as follows:

First, we present two ways of training a neural painter using VAEs and GANs, respectively. We then recreate SPIRAL's CelebA reconstruction results~\cite{ganin2018synthesizing} using a non-reinforcement learning adversarial approach with a neural painter. Concurrent to our work, similar work on training neural painters was done by \citet{zheng2018strokenet} on StrokeNet. Our approach was developed independently at around the same time and uses a different painting program and training methodology, so we believe it can be viewed as a complementary work. We also propose a simple method based on preconditioning to encourage this agent to follow human strokes when reconstructing digits.

In addition, we also propose the use of a neural painter as a differentiable image parameterization~\cite{mordvintsev2018differentiable}. By directly optimizing brushstrokes using backpropagation, we open up a powerful new way to visualize pre-trained image classifiers, by allowing them to “paint” classes they were trained to identify.

Finally, we combine neural painters with neural style transfer~\cite{gatys2015}. By optimizing brushstrokes to minimize only the content loss, we can “paint” the higher-level features of a target image. The artistic medium, in this case, brushstrokes, dictates the style of the resulting image. We call this method \textit{intrinsic style transfer}.

\section{Training Neural Painters}
The main role of a neural painter is to serve as a fully differentiable simulation of a particular painting program. In this paper, we use an open source non-deterministic painting program called MyPaint\cite{mypaint}, which is the same program used by SPIRAL. There are two main considerations for training a neural painter, the painting program's action space and the neural painter's architecture.

\subsection{The Action Space}
The action space defines the set of parameters that are used as control inputs for the painting environment. It serves as the interface that an agent can use to generate a painting.

For the experiments in this paper, we use a slight variation of the action space used by SPIRAL \cite{ganin2018synthesizing}. This action space maps a single action to a single brushstroke in the MyPaint program. An agent “paints” by successively generating actions and applying full brushstrokes on a canvas. 

The action space consists of the following variables:

\begin{itemize}
  \item Start and end pressure - Two variables that define the pressure applied to the brush at the beginning and end of the stroke.
  \item Brush size - Determines the radius of the generated brushstroke.
  \item Color - 3D integer vector determining the RGB color of the brushstroke.
  \item Brush coordinates - Three Cartesian coordinates on a 2D canvas, defining the brushstroke's shape. The coordinates define a starting point, end point, and an intermediate control point, constituting a quadratic Bezier curve.
\end{itemize}

The second consideration in training a neural painter is the architecture. We need an appropriate architecture and training paradigm to learn an accurate mapping from a point in the action space to the corresponding brushstroke. In this paper, we consider 2 approaches for training a neural painter.

\subsection{Training a VAE Neural Painter}

Our first approach is inspired by the two-stage method used by \citet{ha2018recurrent} to learn a world model for a particular environment. \footnote{In the original paper, a VAE \cite{kingma2013auto} is used to learn a latent space for all possible frames in an environment. An RNN is then used to predict the next frame of an environment given an action.}

\begin{figure}[htb!]
  \centering
  \includegraphics[width=12cm, keepaspectratio]{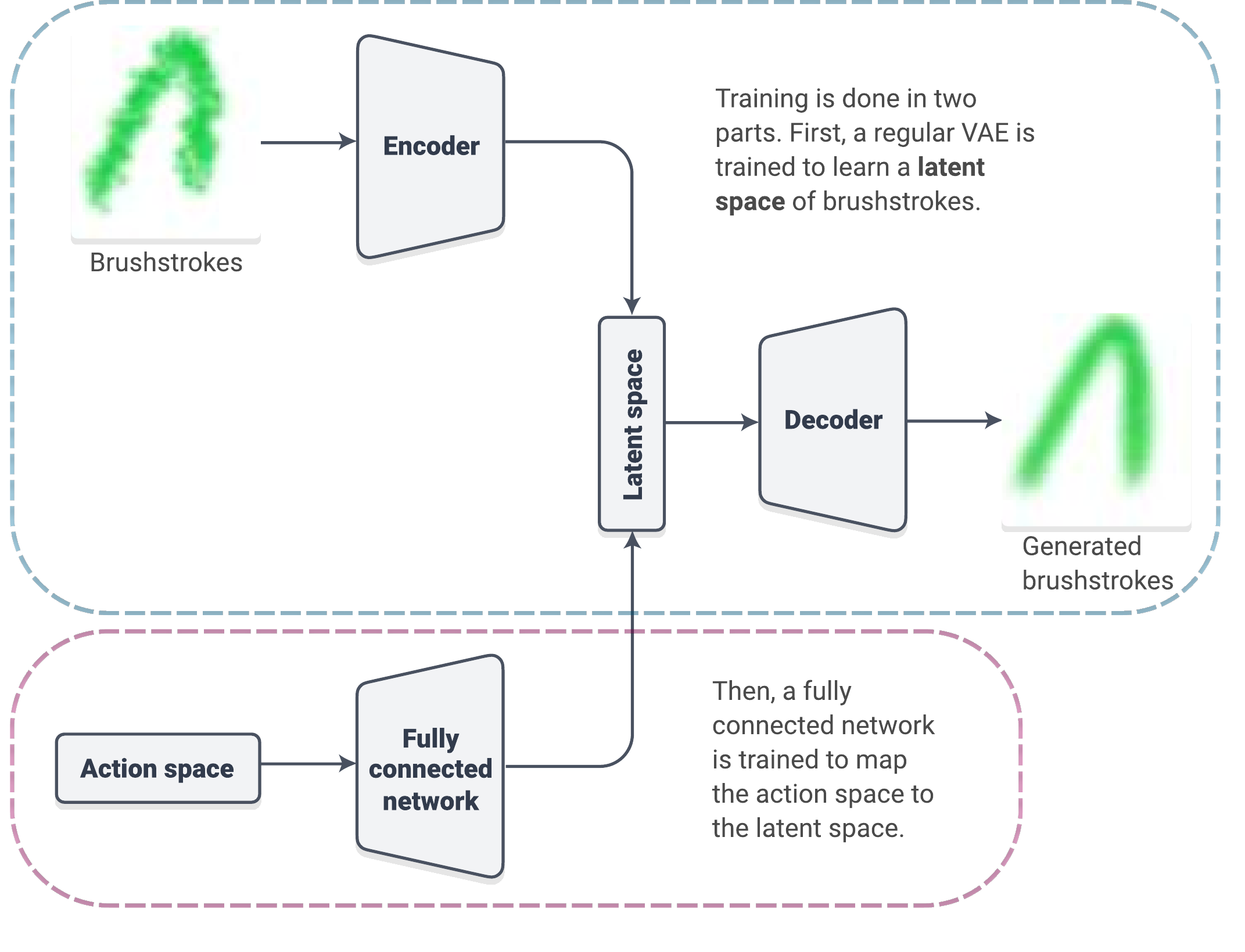}
  \caption{VAE neural painter training process}
  \label{fig:fig1}
\end{figure}

A variational autoencoder (VAE)~\cite{kingma2013auto} is trained to learn a latent space of brushstrokes. We then train a separate network to map an action to the point in latent space corresponding to the expected brushstroke. Unlike the approach in \cite{ha2018recurrent}, we do not need a recurrent neural network (RNN) to map from actions to brushstrokes as there is no relationship between a brushstroke and the previous actions performed on the painting program. Figure \ref{fig:fig1} shows the training process for a VAE neural painter.

\begin{figure}[htb!]
  \centering
  \includegraphics[width=\textwidth, keepaspectratio]{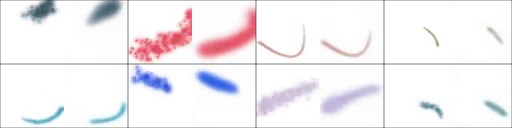}
  \caption{Pairs of real brushstrokes (left) and the corresponding VAE neural painter outputs (right). Notice how the VAE outputs are “smudged” versions of the ground truth.}
  \label{fig:fig2}
\end{figure}

Figure \ref{fig:fig2} compares the results of a trained VAE neural painter with the real output of MyPaint.

The biggest weakness of the VAE neural painter is its “smudging” effect on the brushstrokes. Instead of accurately recreating the dotted texture of the larger brushstrokes, the VAE chooses to smoothen them out instead. Depending on the task, this inaccuracy could lead to less than ideal results when we transfer an agent from a neural painter to the real painting program.

\subsection{Training a GAN Neural Painter}

To solve this problem, we turn to another widely popular family of generative models, generative adversarial networks~\cite{goodfellow2014:adversarial}. GANs have been shown to produce sharper images than VAEs, and this property could help the neural painter produce accurate brushstrokes.

\begin{figure}
  \centering
  \includegraphics[width=12cm, keepaspectratio]{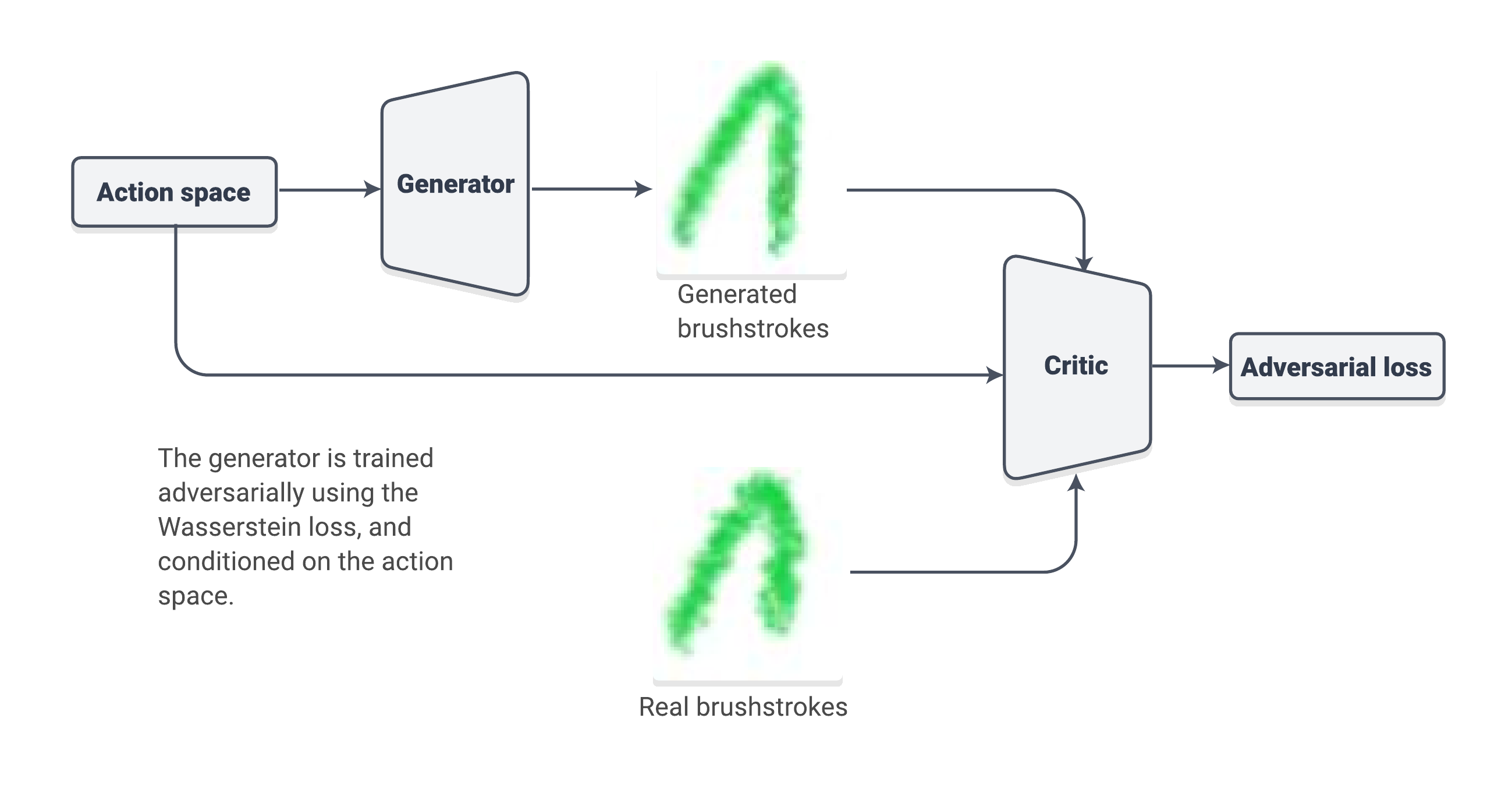}
  \caption{GAN neural painter training process}
  \label{fig:fig3}
\end{figure}

Instead of relying on the reconstruction and KL divergence loss used by VAEs, we use an adversarial loss function to directly learn a mapping from actions to brushstrokes. Unlike a regular GAN, we do not inject noise into the input of the generator. Instead, we feed the generator the input action and have it map directly to a brushstroke. The discriminator is given real and generated action-brushstroke pairs and tries to distinguish whether the pair is valid or not. In this way, it is similar to a conditional GAN~\cite{mirza2014conditional}. The training process, which uses Wasserstein loss~\cite{arjovsky2017wasserstein,gulrajani2017improved}, is illustrated in Figure \ref{fig:fig3}.
 
\begin{figure}
  \centering
  \includegraphics[width=\textwidth, keepaspectratio]{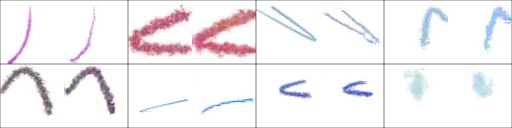}
  \caption{Pairs of real brushstrokes (left) and the corresponding GAN neural painter outputs (right). Notice how the GAN is able to capture the irregularities of real brushstrokes.}
  \label{fig:fig4}
\end{figure}

 The results of training this network is shown in Figure \ref{fig:fig4}. Although the recreation isn't perfect, we can see that the outputs of the neural painter are “rougher” and more realistic as opposed to the smoothed out VAE brushstrokes.
 
How well an agent's actions transfer from a neural painter to the real painting program depend directly on how accurate the neural painter's outputs are. In this section we explored only two possible approaches, and there remains a lot of room for improvement in this area. We have provided accompanying notebooks for training our VAE and GAN neural painters to serve as a good starting point for anyone interested in training their own.

\section{Recreating SPIRAL Results}
\citet{ganin2018synthesizing} trained a neural agent called SPIRAL to learn to “paint” images by using the constrained action space of a real painting program. In the paper, an agent was trained to paint images from three different datasets: MNIST, Omniglot, and CelebA. As the painting program is non-differentiable, the agent was trained using adversarial reinforcement learning.

Since a neural painter is fully differentiable, we don't need reinforcement learning techniques to perform the same experiments. We can simply train the agent using regular adversarial methods.

Our LSTM-based\cite{hochreiter1997long} neural agent is designed to take an input target image and output a set of actions. These output actions are connected directly to the neural painter and mapped to brushstrokes on a canvas. The agent's goal is to recreate the input image on the canvas using the constraints that the neural painter imposes. The idea is that the agent's outputs can be transferred directly to the real painting program, despite seeing only the neural painter during training.

\begin{figure}
  \centering
  \includegraphics[width=12cm, keepaspectratio]{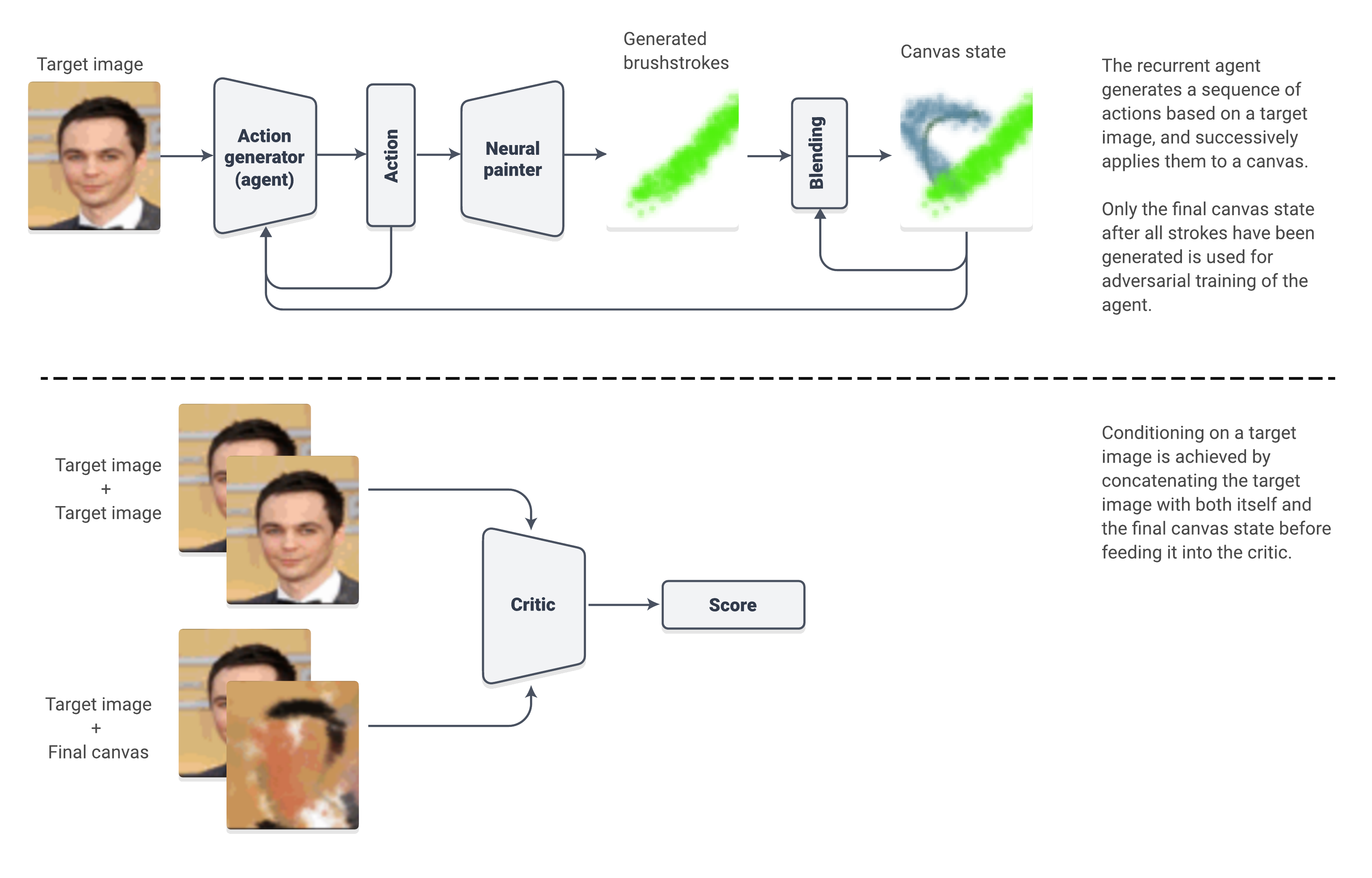}
  \caption{Adversarial training of agent for the purpose of reconstruction}
  \label{fig:fig5}
\end{figure}

When training this agent, instead of directly optimizing a pixelwise loss like L2 distance, we use an adversarial loss. As discussed in the SPIRAL paper, this leads to better and more well-behaved gradients that the agent can use to learn. Figure \ref{fig:fig5} shows the full training setup for our agent.

\begin{figure}
  \centering
  \includegraphics[width=\textwidth, keepaspectratio]{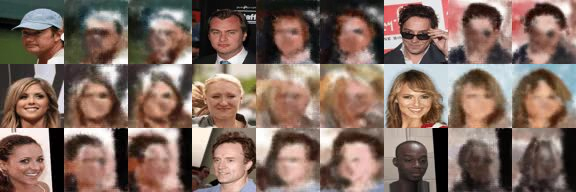}
  \caption{Each group shows three images: the target image (left), the neural painter output (center), and the generated brushstrokes transferred back to MyPaint (right).}
  \label{fig:fig6}
\end{figure}

We test this approach's performance on three datasets: MNIST, KMNIST \cite{clanuwat2018deep}, and CelebA. You can explore the results for our agent in Figure \ref{fig:fig6}.

A significant advantage of this approach over SPIRAL is the amount of computing resources needed to achieve these results. Training SPIRAL involves using several multi-CPU/multi-GPU computers over the course of a few days. Our experiments can be reproduced entirely within the single GPU-environment of Google Colaboratory.

We believe this quick convergence can be partially attributed to the ease of credit assignment. When training our agent, the full gradients from each stroke are available and can be used directly in backpropagation, as opposed to the reinforcement learning paradigm used by SPIRAL, where only the reward at the end of painting is taken into account. This can be observed qualitatively by comparing the strokes produced by these agents on the CelebA dataset. SPIRAL's first few strokes are completely occluded by future strokes, while this does not happen with our approach.

\subsection{The effect of discrete actions}
In the previous section on training neural painters, we mentioned how we removed discrete variables from the action space that SPIRAL used. Why? It is not impossible to train a neural painter on a discrete action space. Our methods for training a neural painter work just as well for a continuous action space as it does with an action space with discrete variables \footnote{Brush size and stroke pressure are discrete variables with 10 levels. An extra binary flag is used to determine whether or not a brush is lifted i.e. a lifted brush produces no stroke.}.

However, there is one very important distinction: neural networks take continuous inputs. Even if a neural painter perfectly recreates the painting program's outputs when given a valid discrete action, its output between two discrete values is completely undefined. The neural painter must bridge this gap, and it is free to do it however it wants.

\begin{figure}
  \centering
  \includegraphics[width=\textwidth, keepaspectratio]{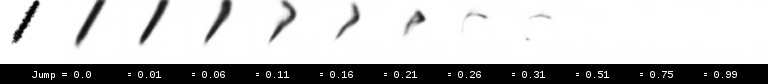}
  \caption{This image shows the effect of lifting the brush partially (i.e. a value between 0 and 1). Values between 0 and 1 are undefined and do not exist in the real environment, however, a neural painter must still “dream” up an output.}
  \label{fig:fig7}
\end{figure}

\begin{figure}
  \centering
  \includegraphics[width=12cm, keepaspectratio]{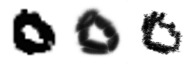}
  \caption{The figure shows three images: the target image (left), the neural painter output (center), and the generated brushstrokes transferred back to MyPaint (right). This result shows the effect of an agent thinking it can use a thicker brush than it actually can.}
  \label{fig:fig8}
\end{figure}

Figure \ref{fig:fig7} shows the output of the neural painter as the lift variable is moved continuously from 0 to 1. It shows an interesting effect. As the brush is lifted, the produced stroke seems to “flicker” until eventually disappearing completely. Somehow, the network has decided that these random flickers were the easiest way to represent a lift value between 0 and 1.

When the goal is to simply recreate a painting program's output, this behavior is not a problem. After all, a user can simply constrain the inputs to use only valid values. Unfortunately, this matters very much when we try to use the gradients from this neural painter to train an agent.

At best, this behavior makes the agent think it can produce impossible strokes, causing a discrepancy between the outputs of the neural painter and the painting program. One specific example is when the neural painter interpolates a stroke thickness between discrete values, which is rounded to the nearest brushstroke size when transferred to the painting program. This is shown in Figure \ref{fig:fig8}.

At worst, it can kill training due to bad gradients. When the lift variable is used, the agent quickly gets into a state where it produces only invisible strokes. At this point, moving the agent's variables slightly in any direction would still produce invisible strokes. Essentially, there is no gradient for the agent to learn anything and training is stuck.

Figuring out how to handle these discrete actions will be an interesting research direction moving forward. Unfortunately, we cannot always side step this issue as we have done in this case by completely ignoring discrete variables. Many interesting environments (including the MuJoCo Scenes environment solved by SPIRAL) will have unavoidable discrete actions, and if we want to apply neural painters to those tasks, handling a discrete action space will be necessary.

\section{Towards Learning Human Strokes}

In the previous section, you may have observed an interesting thing about the stroke order used by the painting agent. For any given agent, the stroke order generated for all target images will be similar.

For example, in the MNIST case, an agent may choose to draw all digits using a bottom-to-top, counter-clockwise approach. The agent does not bother using different strokes for different digits. Notice how 8's are usually treated as 3's with closed loops.

For CelebA, an agent might follow a certain set of steps for every target image e.g. shade the background, fill in the face shape, add hair, then add a stroke for the eyes.

In these experiments, the painting agent was trained with only one goal: recreate a given target image using brushstrokes. Since neural networks are “lazy learners” that converge to the nearest local minimum, the agent learns the simplest possible solution: use similar strokes everywhere. There's no reason to favor dissimilar, let alone human, stroke orders, as long as the final painted image looks as much like the target image as possible.

Of course, there is value in an agent that tries to draw like a human. First, we might be able to learn a model that accurately converts pixel character images to stroke vector data. Second, it might actually improve the original goal of reconstruction. Since digits in the MNIST dataset were actually drawn using a human order, an agent will likely find it easier to reconstruct some of the finer details of an image if it understands how a digit is usually drawn.

In this section, we try a very simple but effective method to bias the agent to learn human strokes: \textit{preconditioning}. Instead of starting adversarial training with the agent's variables initialized randomly, we precondition the agent by forcing it to generate a particular set of strokes for each class. Our process is as follows:

\begin{itemize}
    \item We begin by manually generating a single example for each class, with strokes we think are representative for the entire class. e.g. We can decide to draw 0's counter-clockwise, 1's top-to-bottom.
    \item Train the agent to reconstruct our manual strokes for each class via mean squared error, disregarding adversarial loss. Of course, we do not train the discriminator at this point.
    \item After the agent has started producing our manual strokes for every class, we can consider it preconditioned.
    \item Reintroduce adversarial loss. At this point, we can either completely drop off stroke reconstruction loss, or reduce its effect at a scheduled interval.
    \item Train normally.
\end{itemize}

We test our process on MNIST. Note that the approach requires us to provide only a \textit{single} human example for each class. After training, the agent has more or less learned to stick with our original stroke order, with slight variations to make sure the target image is reconstructed. All 0's are drawn counter-clockwise, and all 1's are drawn top-to-bottom.

One way to explain why preconditioning works is that it changes the basins of attraction for the optimization problem. For an untrained, randomly initialized agent, the nearest local minimum for the adversarial loss is likely one that keeps stroke order similar, regardless of the target character. However, once it has been preconditioned to produce a certain set of strokes for each class, the closest local minimum becomes one that is as similar as possible to those strokes.

Although the approach shows promise, there is much room for improvement. Preconditioning relies on us knowing the correct class label for each image. Without this information, we will not be able to tell the agent to use a certain stroke order for a particular digit, simply because we do not know what digit it is.

Another weakness of the approach is its inability to properly handle multimodal classes that can be written in different ways. An example is the MNIST 7. 7's can be drawn either with or without the horizontal bar at the center, which are two different stroke orders. To apply the same technique, we need to condition the agent in a way that it distinguishes different modes of the class, and have it apply the correct stroke order. Solving these problems are a good direction for future research.

\section{As a Differentiable Image Parameterization}

Differentiable image parameterizations are a technique developed by \citet{mordvintsev2018differentiable} to generate visualizations and art from pre-trained neural networks. Given a network trained on images (usually a convolutional network), we attempt to find a 2D image that maximally activates a particular neuron in the network. Instead of directly optimizing the individual RGB values of each pixel, we try different image generation processes that map some set of parameters to a 2D image. As long as this process is differentiable, we can directly optimize the parameters via backpropagation. Depending on the image generation process, the results can be strikingly different and beautiful. Various parameterizations such as CPPNs~\cite{stanley2007cppn}, Fourier transforms, and 3D to 2D mappings were demonstrated in~\cite{mordvintsev2018differentiable}.

Since neural painters are differentiable mappings from the brushstroke action space to a 2D image, they can be used directly as a differentiable image parameterization.

\subsection{Visualizing ImageNet classes}

\begin{figure}[!htb]
  \centering
  \includegraphics[width=12cm, keepaspectratio]{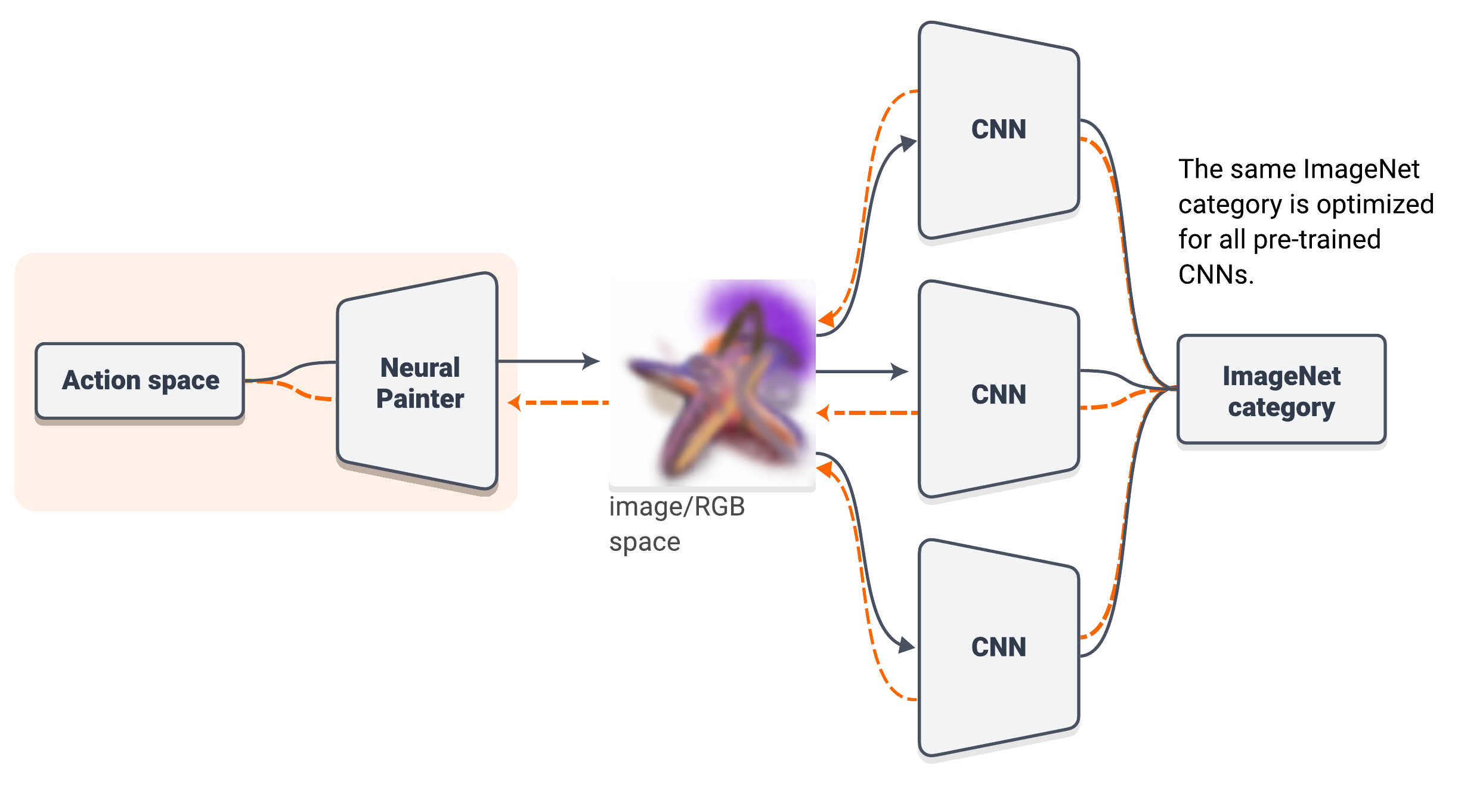}
  \caption{A neural painter as a differentiable image parameterization.}
  \label{fig:fig9}
\end{figure}

We focus on visualizing the final layer of the pre-trained networks, corresponding to specific ImageNet classes. We find that a good way to improve the outputs is to optimize more than one pre-trained network at the same time\footnote{As done by Tom White in his work on Synthetic Abstractions \cite{white2018}}. This helps the outputs generalize better by reducing the effect of each individual network's imperfections. Figure \ref{fig:fig9} shows how to use a neural painter as a differentiable image parameterization to visualize ImageNet classes.

\begin{figure}[!htb]
  \centering
  \includegraphics[width=11cm, keepaspectratio]{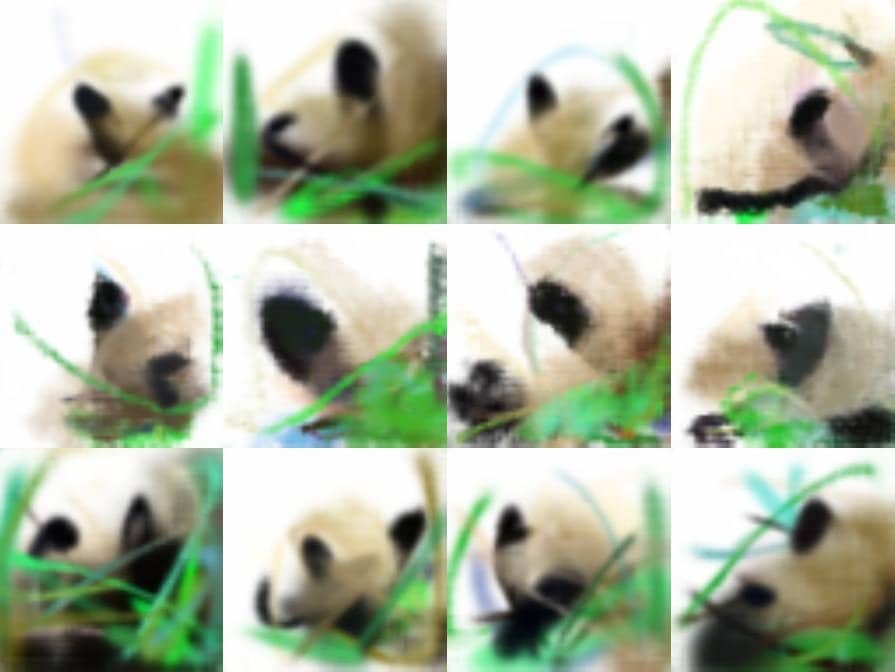}
  \caption{A neural network's paintings of the “optimal” pandas}
  \label{fig:fig10}
\vskip -0.15 in
\end{figure}

A fun way to interpret the results of a neural painter used as a differentiable image parameterization is as the answer to the question:

\begin{center}
\textit{If you gave a pre-trained network a brush and asked it to paint a picture of the optimal panda, what~would~it~paint?}
\end{center}

Figure \ref{fig:fig10} shows examples of outputs generated by optimizing different ImageNet classes. The results show just how diverse the generated outputs can be for any given class, by simply tweaking the number of strokes, changing the neural painter, or using different pre-trained networks.

\subsection{Intrinsic Style Transfer}

\begin{figure}[!htb]
  \centering
  \includegraphics[width=12cm, keepaspectratio]{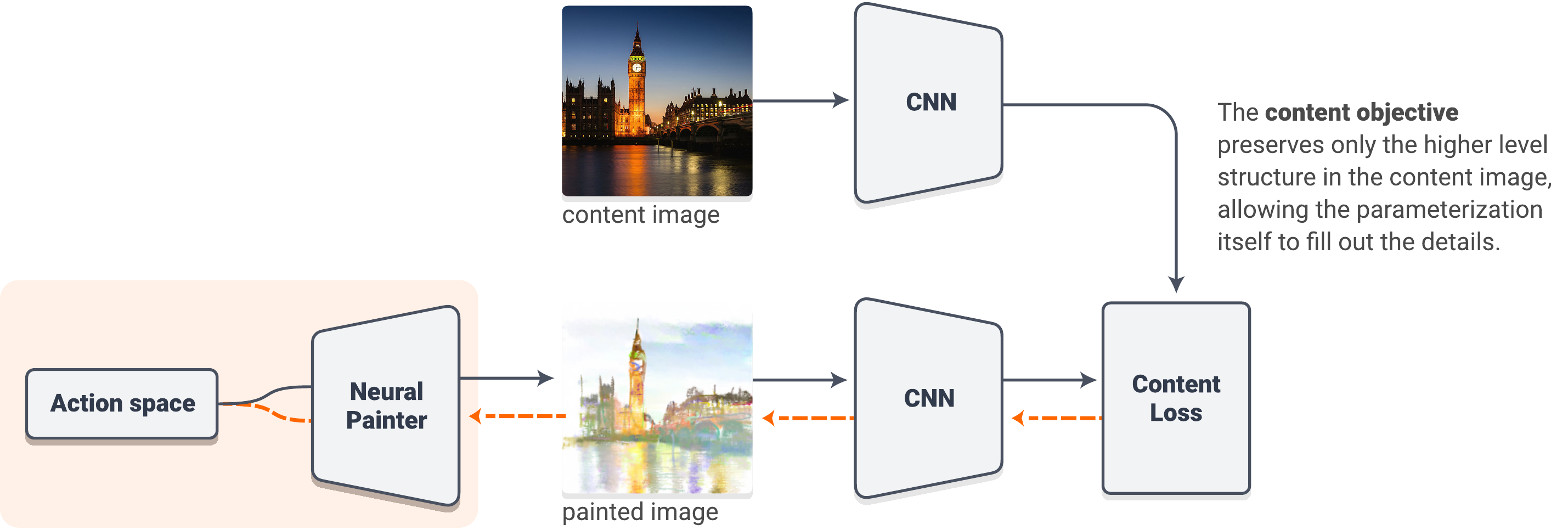}
  \caption{Intrinsic style transfer using a neural painter}
  \label{fig:fig11}
\vskip -0.1 in
\end{figure}

As a differentiable image parameterization, neural painters are not limited to visualizing layers of a pre-trained neural network. We can produce various interesting effects depending on the loss we are optimizing for. One such effect is stroke-based painterly rendering~\cite{hertzmann2003survey} of a target image. With this method, we optimize brushstrokes to minimize only the content loss\footnote{To calculate the content loss between a content image and an output image, we take the mean squared error of their respective activations for a particular layer in a pre-trained neural network.} in neural style transfer~\cite{gatys2015}. This setup is shown in Figure \ref{fig:fig11}.

\begin{figure}
  \centering
  \includegraphics[width=\textwidth, keepaspectratio]{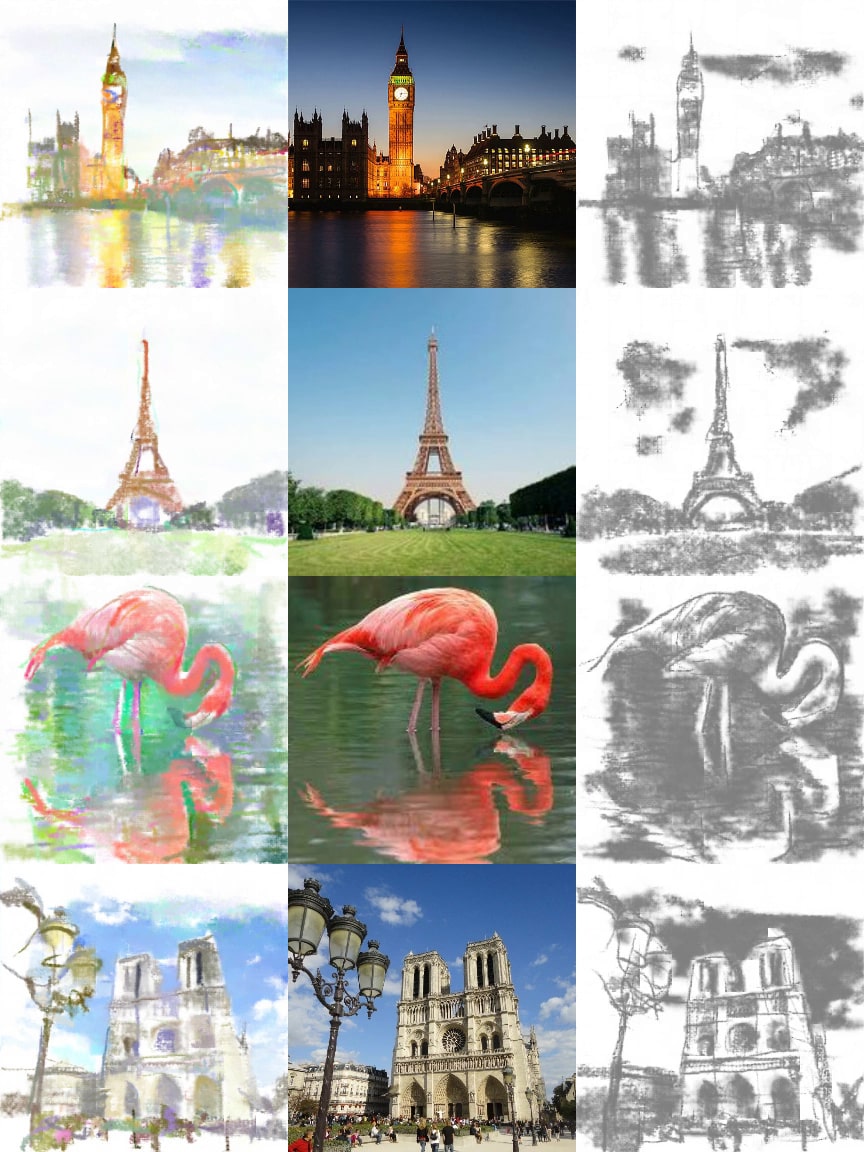}
  \caption{Results for intrinsic style transfer using a GAN neural painter for both colored and grayscale brushstrokes. The strokes are generated on multiple overlapping grids. Although the neural painter is only designed to output 64x64 pixels on a canvas, we can stitch multiple canvases together to achieve an arbitrary resolution, limited only by GPU memory.}
  \label{fig:fig13}
\end{figure}

Intuitively, this technique lets us find brushstrokes that preserve only the higher-level content in the target image. The effect produced is that of painting only the meaningful parts of a target image, without caring about pixel-level reconstruction. The \textit{style} is an intrinsic property dictated purely by the artistic medium, in this case, brushstrokes. Figure \ref{fig:fig13} shows some results of intrinsic style transfer.

By manually changing the primitives of the brushstroke, we can achieve vastly different styles. Note the difference in outputs by simply constraining the brush to use only grayscale values. Finding new styles by applying different constraints or using different artistic mediums\footnote{Constraints could be simple modifications such as changing the color palette or making brushstrokes thinner. A good place to start looking for interesting brushstroke constraints is the extensive literature available on stroke-based rendering~\cite{hertzmann2003survey}. We can also completely change the medium and try differentiable parameterizations like CPPNs, which have been compared to light paintings\cite{mordvintsev2018differentiable}.} will be an exciting research path forward.

\section{Conclusions}

Constraints are a key element of creativity. The natural constraints that an artistic medium imposes upon the artist give a piece of art a distinct look from others. An artist attempting to paint a scene using oil paints will get a remarkably different result from someone trying to sketch the same scene with a pencil. In the same sense, using a neural painter leads to creative ways to achieve an objective - whether it be recreating digits, painting faces, or maximizing a pre-trained network's activations.

This paper explored the power of neural painters - a differentiable constraint learned from a non-differentiable real-life constraint. We believe this concept could be extended to different artistic mediums, such as splatter painting, or even 3D sculptures.

\subsubsection*{Acknowledgments}

We would like to thank David Ha for his detailed feedback and encouragement throughout this work. We are also grateful to Ludwig Schubert and the Distill community for providing support for the diagrams in this paper.

Many of our diagrams were repurposed from the article Differentiable Image Parameterizations by \citet{mordvintsev2018differentiable}

\begin{figure}
  \centering
  \includegraphics[width=12cm, keepaspectratio]{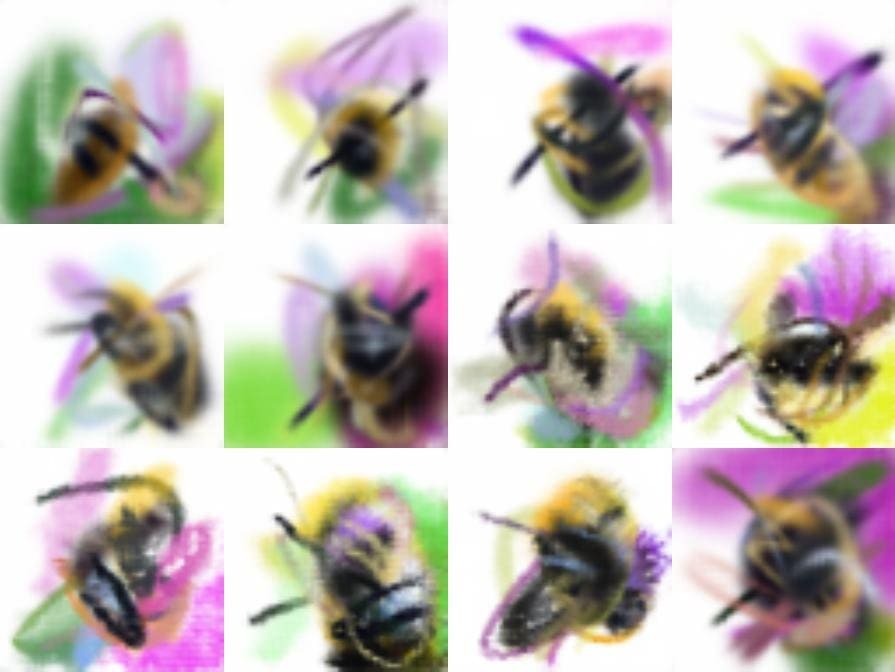}
  \caption{A neural network's paintings of the \textit{optimal} bees}
  \label{fig:fig14}
\end{figure}

\begin{figure}
  \centering
  \includegraphics[width=12cm, keepaspectratio]{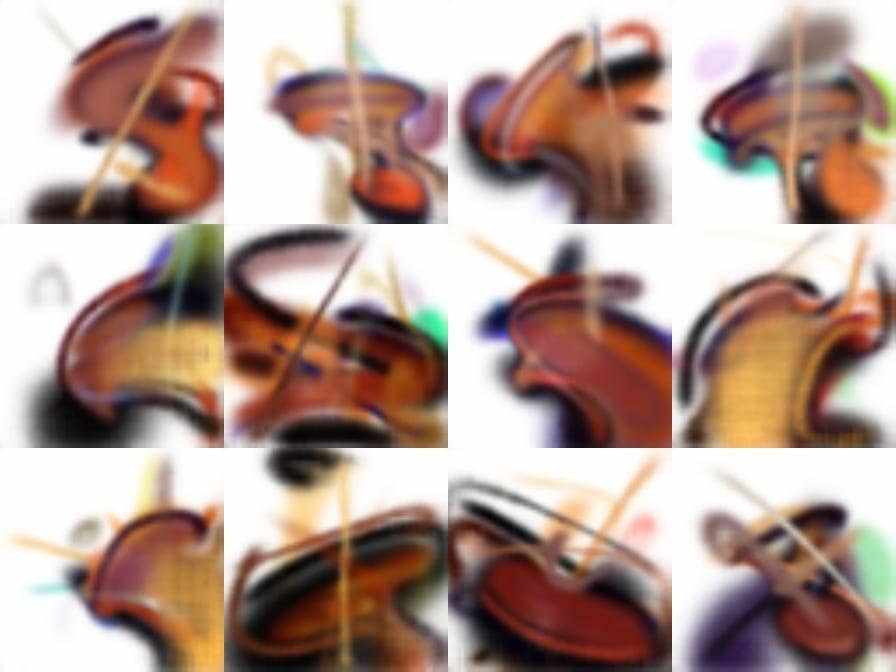}
  \caption{A neural network's paintings of the \textit{optimal} violins}
  \label{fig:fig15}
\end{figure}

\begin{figure}
  \centering
  \includegraphics[width=12cm, keepaspectratio]{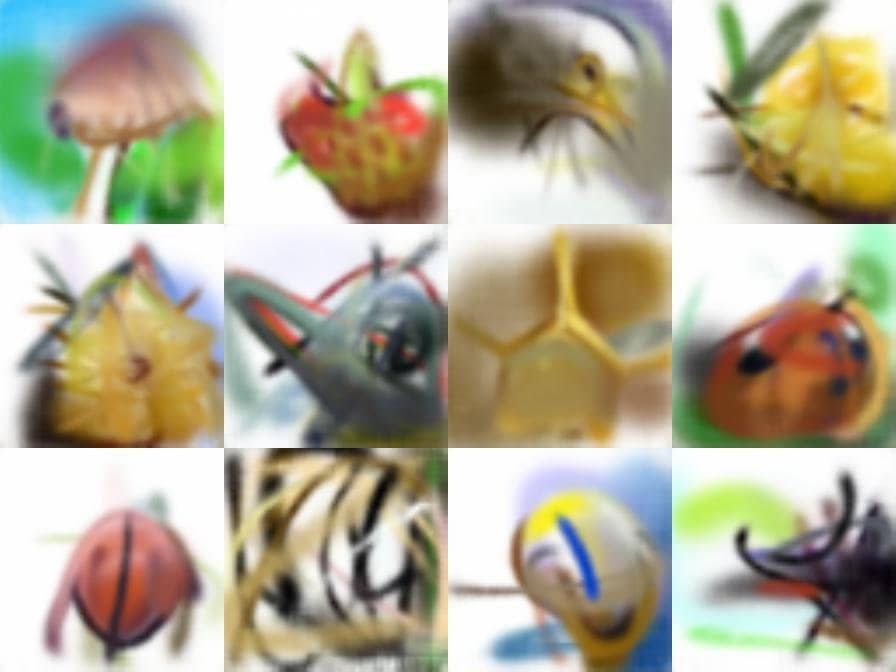}
  \caption{Other ImageNet class visualizations}
  \label{fig:fig16}
\end{figure}

\clearpage
\bibliographystyle{plainnat}  
\bibliography{references}  

\begin{thebibliography}{27}
\providecommand{\natexlab}[1]{#1}
\providecommand{\url}[1]{\texttt{#1}}
\expandafter\ifx\csname urlstyle\endcsname\relax
  \providecommand{\doi}[1]{doi: #1}\else
  \providecommand{\doi}{doi: \begingroup \urlstyle{rm}\Url}\fi

\bibitem[Arjovsky et~al.(2017)Arjovsky, Chintala, and
  Bottou]{arjovsky2017wasserstein}
Martin Arjovsky, Soumith Chintala, and L{\'e}on Bottou.
\newblock Wasserstein gan.
\newblock \emph{arXiv preprint arXiv:1701.07875}, 2017.

\bibitem[Clanuwat et~al.(2018)Clanuwat, Bober-Irizar, Kitamoto, Lamb, Yamamoto,
  and Ha]{clanuwat2018deep}
Tarin Clanuwat, Mikel Bober-Irizar, Asanobu Kitamoto, Alex Lamb, Kazuaki
  Yamamoto, and David Ha.
\newblock Deep learning for classical japanese literature.
\newblock \emph{arXiv preprint arXiv:1812.01718}, 2018.

\bibitem[contributors(2019)]{mypaint}
MyPaint contributors.
\newblock Mypaint, 2019.
\newblock URL \url{https://github.com/mypaint/mypaint}.

\bibitem[Elgammal et~al.(2017)Elgammal, Liu, Elhoseiny, and
  Mazzone]{elgammal2017can}
Ahmed Elgammal, Bingchen Liu, Mohamed Elhoseiny, and Marian Mazzone.
\newblock Can: Creative adversarial networks, generating" art" by learning
  about styles and deviating from style norms.
\newblock \emph{arXiv preprint arXiv:1706.07068}, 2017.

\bibitem[Frans and Cheng(2018)]{frans2018unsupervised}
Kevin Frans and Chin-Yi Cheng.
\newblock Unsupervised image to sequence translation with canvas-drawer
  networks.
\newblock \emph{arXiv preprint arXiv:1809.08340}, 2018.

\bibitem[Ganin et~al.(2018)Ganin, Kulkarni, Babuschkin, Eslami, and
  Vinyals]{ganin2018synthesizing}
Yaroslav Ganin, Tejas Kulkarni, Igor Babuschkin, SM~Eslami, and Oriol Vinyals.
\newblock Synthesizing programs for images using reinforced adversarial
  learning.
\newblock \emph{arXiv preprint arXiv:1804.01118}, 2018.

\bibitem[Gatys et~al.(2015)Gatys, Ecker, and Bethge]{gatys2015}
Leon~A. Gatys, Alexander~S. Ecker, and Matthias Bethge.
\newblock A neural algorithm of artistic style.
\newblock \emph{CoRR}, abs/1508.06576, 2015.
\newblock URL \url{http://arxiv.org/abs/1508.06576}.

\bibitem[Gatys et~al.(2016)Gatys, Ecker, and Bethge]{Gatys_2016_CVPR}
Leon~A. Gatys, Alexander~S. Ecker, and Matthias Bethge.
\newblock Image style transfer using convolutional neural networks.
\newblock In \emph{The IEEE Conference on Computer Vision and Pattern
  Recognition (CVPR)}, June 2016.

\bibitem[Goodfellow et~al.(2014)Goodfellow, Pouget-Abadie, Mirza, Xu,
  Warde-Farley, Ozair, Courville, and Bengio]{goodfellow2014:adversarial}
Ian Goodfellow, Jean Pouget-Abadie, Mehdi Mirza, Bing Xu, David Warde-Farley,
  Sherjil Ozair, Aaron Courville, and Yoshua Bengio.
\newblock Generative adversarial nets.
\newblock In Z.~Ghahramani, M.~Welling, C.~Cortes, N.~D. Lawrence, and K.~Q.
  Weinberger, editors, \emph{Advances in Neural Information Processing Systems
  27}, pages 2672--2680. Curran Associates, Inc., 2014.
\newblock URL
  \url{http://papers.nips.cc/paper/5423-generative-adversarial-nets.pdf}.

\bibitem[Gulrajani et~al.(2017)Gulrajani, Ahmed, Arjovsky, Dumoulin, and
  Courville]{gulrajani2017improved}
Ishaan Gulrajani, Faruk Ahmed, Martin Arjovsky, Vincent Dumoulin, and Aaron~C
  Courville.
\newblock Improved training of wasserstein gans.
\newblock In \emph{Advances in Neural Information Processing Systems}, pages
  5767--5777, 2017.

\bibitem[Ha and Schmidhuber(2018)]{ha2018recurrent}
David Ha and J{\"u}rgen Schmidhuber.
\newblock Recurrent world models facilitate policy evolution.
\newblock In \emph{Advances in Neural Information Processing Systems}, pages
  2450--2462, 2018.
\newblock URL \url{https://worldmodels.github.io/}.

\bibitem[Haeberli(1990)]{haeberli1990paint}
Paul Haeberli.
\newblock Paint by numbers: Abstract image representations.
\newblock In \emph{ACM SIGGRAPH computer graphics}, volume~24, pages 207--214.
  ACM, 1990.

\bibitem[Hafner et~al.(2018)Hafner, Lillicrap, Fischer, Villegas, Ha, Lee, and
  Davidson]{hafner2018learning}
Danijar Hafner, Timothy Lillicrap, Ian Fischer, Ruben Villegas, David Ha,
  Honglak Lee, and James Davidson.
\newblock Learning latent dynamics for planning from pixels.
\newblock \emph{arXiv preprint arXiv:1811.04551}, 2018.

\bibitem[Hertzmann(1998)]{hertzmann1998painterly}
Aaron Hertzmann.
\newblock Painterly rendering with curved brush strokes of multiple sizes.
\newblock In \emph{Proceedings of the 25th annual conference on Computer
  graphics and interactive techniques}, pages 453--460. ACM, 1998.

\bibitem[Hertzmann(2003)]{hertzmann2003survey}
Aaron Hertzmann.
\newblock A survey of stroke-based rendering.
\newblock \emph{IEEE Computer Graphics and Applications}, \penalty0
  (4):\penalty0 70--81, 2003.

\bibitem[Hochreiter and Schmidhuber(1997)]{hochreiter1997long}
Sepp Hochreiter and J{\"u}rgen Schmidhuber.
\newblock Long short-term memory.
\newblock \emph{Neural computation}, 9\penalty0 (8):\penalty0 1735--1780, 1997.

\bibitem[Jing et~al.(2017)Jing, Yang, Feng, Ye, Yu, and Song]{jing2017neural}
Yongcheng Jing, Yezhou Yang, Zunlei Feng, Jingwen Ye, Yizhou Yu, and Mingli
  Song.
\newblock Neural style transfer: A review.
\newblock \emph{arXiv preprint arXiv:1705.04058}, 2017.

\bibitem[Kingma and Welling(2013)]{kingma2013auto}
Diederik~P Kingma and Max Welling.
\newblock Auto-encoding variational bayes.
\newblock \emph{arXiv preprint arXiv:1312.6114}, 2013.

\bibitem[LeCun et~al.(2015)LeCun, Bengio, and Hinton]{lecun2015deep}
Yann LeCun, Yoshua Bengio, and Geoffrey Hinton.
\newblock Deep learning.
\newblock \emph{nature}, 521\penalty0 (7553):\penalty0 436, 2015.

\bibitem[Mirza and Osindero(2014)]{mirza2014conditional}
Mehdi Mirza and Simon Osindero.
\newblock Conditional generative adversarial nets.
\newblock \emph{arXiv preprint arXiv:1411.1784}, 2014.

\bibitem[Mordvintsev et~al.(2018)Mordvintsev, Pezzotti, Schubert, and
  Olah]{mordvintsev2018differentiable}
Alexander Mordvintsev, Nicola Pezzotti, Ludwig Schubert, and Chris Olah.
\newblock Differentiable image parameterizations.
\newblock \emph{Distill}, 3\penalty0 (7):\penalty0 e12, 2018.

\bibitem[Shiraishi and Yamaguchi(2000)]{shiraishi2000algorithm}
Michio Shiraishi and Yasushi Yamaguchi.
\newblock An algorithm for automatic painterly rendering based on local source
  image approximation.
\newblock In \emph{NPAR}, pages 53--58. Citeseer, 2000.

\bibitem[Stanley(2007)]{stanley2007cppn}
Kenneth~O Stanley.
\newblock Compositional pattern producing networks: A novel abstraction of
  development.
\newblock \emph{Genetic programming and evolvable machines}, 8\penalty0
  (2):\penalty0 131--162, 2007.
\newblock URL \url{http://eplex.cs.ucf.edu/papers/stanley_gpem07.pdf}.

\bibitem[White(2018)]{white2018}
Tom White.
\newblock Perception engines, 2018.
\newblock URL
  \url{https://medium.com/artists-and-machine-intelligence/perception-engines-8a46bc598d57}.

\bibitem[Winkenbach and Salesin(1994)]{winkenbach1994computer}
Georges Winkenbach and David~H Salesin.
\newblock Computer-generated pen-and-ink illustration.
\newblock In \emph{Proceedings of the 21st annual conference on Computer
  graphics and interactive techniques}, pages 91--100. ACM, 1994.

\bibitem[Xie et~al.(2013)Xie, Hachiya, and Sugiyama]{xie2013artist}
Ning Xie, Hirotaka Hachiya, and Masashi Sugiyama.
\newblock Artist agent: A reinforcement learning approach to automatic stroke
  generation in oriental ink painting.
\newblock \emph{IEICE TRANSACTIONS on Information and Systems}, 96\penalty0
  (5):\penalty0 1134--1144, 2013.

\bibitem[Zheng et~al.(2019)Zheng, Jiang, and Huang]{zheng2018strokenet}
Ningyuan Zheng, Yifan Jiang, and Dingjiang Huang.
\newblock Strokenet: A neural painting environment.
\newblock In \emph{International Conference on Learning Representations
  (ICLR)}, 2019.

\end{thebibliography}


\end{document}